# Identifying roles of clinical pharmacy with survey evaluation

Andreja Čufar, University Medical Centre Ljubljana, Pharmacy, Zaloška 7, 1000 Ljubljana

Aleš Mrhar, University of Ljubljana, Faculty of Pharmacy, Aškerčeva 7, 1000 Ljubljana

Marko Robnik-Šikonja, University of Ljubljana, Faculty of Computer and Information Science, Tržaška 25, 1000 Ljubljana (corresponding author: marko.robnik@fri.uni-lj.si)

**Abstract**

The survey data sets are important sources of data and their successful exploitation is of key importance for informed policy-decision making. We present how a survey analysis approach initially developed for customer satisfaction research in marketing can be adapted for the introduction of clinical pharmacy services into hospital. We use two analytical approaches to extract relevant managerial consequences. With OrdEval algorithm we first evaluate the importance of competences for the users of clinical pharmacy and extract their nature according to the users' expectations. Next, we build a model for predicting a successful introduction of clinical pharmacy to the clinical departments. We the wards with the highest probability of successful cooperation with a clinical pharmacist. We obtain useful managerially relevant information from a relatively small sample of highly relevant respondents. We show how the OrdEval algorithm exploits the information hidden in the ordering of class and attribute values and their inherent correlation. Its output can be effectively visualized and complemented with confidence intervals.

Keyword: data mining, survey analysis, classification, clinical pharmacy, HR management

# 1   Introduction

In spite of a plethora of new information sources available, surveys are still the most important and frequently used tool for managers and decision makers when investigating alternative outcomes and possible strategies before making important decisions. Any progress made in this standard tool which allows better insight into the information gathered is therefore an important contribution, not only to the field of data analytics, but also to the applicative fields using the surveys.

Clinical pharmacy is a relatively new discipline in the pharmacy profession which is patient rather than drug oriented and aims to improve the quality of drug therapy. The Clinical pharmacist's activity is therefore patient cantered, cooperative, and inter-professional as they have to collaborate with physicians and nurses in the health care team (Hepler, 2004; Miller, 1981). The added value of clinical pharmacy can be proven based on numerous publications (Bond & Raehl, 2008; Borenstein et al., 2003; Phillips & Lipman, 1981). Despite many positive examples of cooperation between physicians and pharmacists, there are still some obstacles (De Rijdt, Willems, & Simoens, 2008) hindering the systemic regulation of clinical pharmacy services, which would promote this cooperation more widely. The physicians' fear of losing independence and professional autonomy is one of the barriers and reasons for the negative attitude towards clinical pharmacists as inspectors of their work (Alkhateeb et al., 2009). Physicians who have already had an opportunity to cooperate with a clinical pharmacist are more open to this cooperation than those who lack such kind of an experience (Chevalier & Neville, 2011; Rauch, 1982; Zillich et al., 2004). Physicians' and nurses' acceptance of the clinical pharmacist depends on how well their activities meet their needs and expectations. Physicians' perceptions of the importance of clinical pharmacy activities differ substantially from pharmacists' perceptions (Sulick & Pathak, 1996). Therefore it is necessary to recognize the most important clinical pharmacy activities from the perspective of its users, namely physicians' and nurses'.

According to the behavioral decision theory and total quality management approach of Kano (Berger et al., 1993; Kano et al. 1984), the analysis of customer needs is a three phase process: (1) understanding customer preferences, (2) requirement prioritization, and (3) requirement classification. The Kano model is widely used in several industries as an effective tool to understanding customer preferences. Kano proposed a two dimensional system for quality management, where he identified three levels of customer expectations. Satisfying basic expectations keeps customer satisfied, while not meeting his expectations makes him dissatisfied. He denoted this kind of expectations as dissatisfies or basic or must be. The second type of expectations make customer satisfied/dissatisfied depending on the quality/quantity of performance, that's why he called them performance expectations. The third level of customer's expectations is hidden, not self-confident expectations. Satisfying those makes customer delighted. These are delighters or excitement expectations.

We present how a survey analysis approach initially developed for customer satisfaction research in marketing can be adapted to a completely different problem, namely the introduction of clinical pharmacy services into hospital. The approach is based on the evaluation of ordinal attributes i.e. survey questions and their relation to the expected outcome. By taking the ordinal nature of many surveys' questions into account the approach allows a sort of what-if analysis where probabilities of the expected outcome are computed conditioned on changes in outcome of survey questions. An example of the output would be a probability of more successful service introduction if certain feature of the service is better communicated to the users. As the human resources are limited, such knowledge is valuable for the management who wants to allocate the resources efficiently. The proposed approach also allows a categorization of attributes according to the Kano model.

The OrdEval algorithm (Robnik-Šikonja, Brijs, & Vanhoof, 2009; Robnik-Šikonja & Vanhoof, 2007) is an analytical tool for evaluation of the importance and impact of various factors in the given survey. In the analysis of customer satisfaction data for a particular product/service, OrdEval can determine the importance of each product's feature to the overall customer's satisfaction, and also indicate the thresholds where satisfaction with individual feature starts having strong positive or negative impact on the overall satisfaction. The outputs of OrdEval are probabilistic factors indicating the probability that an increase/decrease in the individual feature or the feature's value will have impact on the dependent variable. The intuition behind this approach is to approximate the inner workings of the decision process taking place in each individual respondent, which forms a relationship between the features and the response. If such introspection would be possible, one could observe a causal effect the change of a feature's value has on the response value. By measuring such an effect we could reason about the importance of the feature's values and the type of the attribute. Also, we could determine which values are thresholds for a change of behavior. OrdEval algorithm uses the data sample and approximates this reasoning. For each respondent it selects respondents most similar to it, and does inference based on them. For example, to evaluate the effect an increase in certain feature value would have on overall satisfaction, the algorithm computes the probability for such an effect from the similar respondents with increased value of that feature. To get statistically valid and practically interesting results, the overall process is repeated for a large enough number of respondents, and weighted with a large enough number of similar respondents. The motivation and contribution of this paper is to demonstrate how OrdEval works in a medical management context, how its output can be visualized and adapted to include information relevant for decision markers, and the new insights into clinical pharmacy services, which is used as our application topic.

With this research we aim to demonstrate the use of survey analysis method OrdEval in a novel context of clinical pharmacy, where feature evaluation and detected value thresholds identify the key activities needed to satisfy the expectations of physicians and nurses. This information coupled with the prediction model for introduction of clinical pharmacy services in the hospital is a practically useful aid for decision making.

The paper is divided into five sections. In Section 2 we present clinical pharmacy and the activities it encompasses, followed by the survey we designed to measure the expectations of doctors, nurses, and pharmacists. In Section 3 we describe OrdEval approach to evaluation of attributes and the way the attributes can be interpreted. In Section 4 we present the results of the analysis with the emphasis on the methodological aspects of OrdEval. In Section 5 we conclude with the overview of the contributions and plans for further work.

## 2  Collecting clinical pharmacy data

University Medical Centre (UMC) Ljubljana is the largest health care institution and university hospital in Slovenia. It has nearly 2200 beds and annually over 100,000 hospitalizations, more than 750,000 ambulatory visits and over 250,000 functional diagnostics visits. On December 31 2013 there were employed nearly 1200 physicians and 3.750 nurses. The hospital has a central pharmacy with 45 pharmacists (December 31, 2013). The average cost of medications is approximately 350€ per patient. As one of the main strategic goals of the hospital pharmacy department, a stepwise introduction of clinical pharmacy service in the hospital began in 2010 with a strong support of the top management. There are 64 clinics, clinical departments, centers and institutes operating within the hospital. 49 of them are using medical preparations while performing their practice. 6 out of 49 departments have negligible drug consumption, so clinical pharmacy service would be useless there. In the other 43 departments there is a considerable amount of medications used; therefore we can expect clinical pharmacy service being valuable.

Clinical pharmacy service is a set of activities including, but not limited to, providing and validation of all kinds of information regarding medicines, cooperation in planning and performing of pharmacotherapy, medication reconciliation, verification of prescribing, preparation and administration of medicines, outcomes assessment, procurement of quality medicines and safe and cost effective medicines utilization assurance.

A comprehensive literature search was performed to prepare a list of all possible clinical pharmacy activities. The list was further used to construct a survey questionnaire with Likert measurement scale to conduct a descriptive observational attitude study of physicians' and nurses' opinion about the importance of each of the listed activities and competencies of clinical pharmacists. The participants had to choose the level of agreement with each of the listed affirmative statements in the questionnaire. We decided to perform the study on the basis of directed sampling only among management staff members because we were interested in getting the middle management's opinion which we expected to have a higher managerial impact than a wider population sampling among other staff members with less decision making influence. Consequently the questionnaire was sent only to medical directors and heads of departments – and to their head nurses. They were allowed to share the questionnaire to their colleagues – lover level leading staff, upon their judgment.

The questionnaire was composed of three types of questions. In the first part of the questionnaire (17 questions) clinical pharmacy activities pertaining to the hospital system were stated, while the second part of the questionnaire (19 questions) contained activities directly connected to an individual patient care. The third part (16 questions) of the questionnaire dealt with clinical pharmacist's competencies. The participants had to choose the level of agreement on the Likert scale from 1(I totally disagree) to 5 (I totally agree) with each of the listed affirmative statements in the parts one and two of the questionnaire while indicating the importance (form least important – 1 to very important 5) of a particular competence in the third part. The questionnaire was first tested on a group of pharmacists to check the scope of selected clinical pharmacy activities. In the second step it was validated by a group of experts: a physician, a non-hospital pharmacist and an independent human resource manager.

The questionnaire was sent to 43 physicians – medical directors or heads of departments – and to their head nurses. Beside them, the questionnaire was also sent by e-mail to all pharmacists employed in the hospital pharmacy at the time of performing the survey (27). The pharmacists were instructed to answer the questionnaire only in case they were interested in participating in the performing of clinical pharmacy services. The participants could choose to fill the questionnaire electronically or to print it out and fill it in manually. 43 physicians from 27 clinics/clinical departments and 42 nurses from 26 clinics/clinical departments filled in the questionnaire. Some of the chief physicians and chief nurses distributed the questionaries' also to heads of sub-departments, while on some other departments, physicians or nurses did not participate. Total number of participating physicians and nurses from clinics – divisions of University Medical Centre Ljubljana is presented in table 1.

**Table 1: Total number of participating physicians and nurses from clinics – divisions of University Medical Centre Ljubljana**

| CLINIC | PHYSICIANS | NURSES |
|---|---|---|
| INTERNAL | 14 | 12 |
| SURGICAL | 1 | 10 |
| NEUROLOGY | 3 | 11 |
| GYNECOLOGY | 5 | 1 |
| PEDIATRIC | 11 | 2 |
| FOR INFECTIOUS DISEASES AND FEBRILE ILLNESSES | 7 | 3 |
| OPHTALMOLOGY | 1 | 1 |
| DERMATOVENEROLOGY | 1 | 0 |
| ORTHOPAEDIC | 0 | 1 |
| OTORHINOLARYNGOLOGY AND CERVICOFACIAL SURGERY | 0 | 1 |

Out of total 27 pharmacists, 13 expressed their wish to work as clinical pharmacists and filled in the questionnaire. The response of clinical pharmacists was therefore considered as 100%. 82 responses from physicians and nurses as users of clinical pharmacy service were analyzed in comparison to 13 responses of pharmacists as providers of the service. The survey results were collected in a tabular form, together with the demographic data of each respondent: profession (physician, nurse or pharmacist), name of the clinic/clinical department, age and gender, and the level of agreement (from 1 to 5) with each statement about clinical pharmacy services. In the second stage of our research we established a collaboration of clinical pharmacists on some wards providing a selected range of clinical pharmacy activities. We selected activities most favored by physicians and nurses and for which pharmacists felt they have competences to provide them. We used a spreadsheet to keep records about the type of clinical pharmacy activities performed and time spent for each of them.

For the wards cooperating with a clinical pharmacist an estimated satisfaction score and an estimated influence of particular activity on the total satisfaction with the cooperation was assigned by the head of the pharmacy.

The collected data was than divided into two groups, the first group of data being collected from participants already cooperating with clinical pharmacist to serve as a training data set, and the other collected from participants lacking this cooperation forming a prediction data set.

# 3   Analytical approach

Machine learning and data mining have been used extensively in medicine and pharmacology. In a recent review (Esfandiary et al., 2014) medical data mining is divided to six medical tasks (screening, diagnosis, treatment, prognosis, monitoring and management) and for each task five data mining approaches are studied (classification, regression, clustering, association and hybrid). Our work could be classified as management task, but the methodology presented is aimed towards decision support, which is also frequently used in pharmacology, e.g. (Rommers, 2013). Our approach is based on evaluation of ordinal features, though we use also classification. This is different to classical and modern approaches to evaluation of services and also different to classical analysis of surveys in medical decision support, e.g. (Scheepers-Hoeks, 2013).

We use two analytical approaches to extract relevant managerial consequences. With OrdEval algorithm we first evaluate the importance of competences for the users of clinical pharmacy and extract their type according to the users' expectations. Next, we build a model for predicting a successful introduction of clinical pharmacy to the clinical departments.

## 3.1  Evaluation of ordinal features

Based on the Kano model of customer satisfaction we aim to classify clinical pharmacy activities into three categories of attributes, namely basic or threshold, performance, and excitement activities. Not meeting the expectations of basic activities would cause dissatisfaction with the new clinical pharmacy service. The influence of performance activities on the total value of clinical pharmacy is approximately linear, while excitement activities have a positive impact on the satisfaction with clinical pharmacy services when provided, but do not affect the satisfaction value when not provided and can therefore be an opportunity to increase the overall satisfaction (Deng et al., 2008). Figure 1 presents attributes' characteristics according to the Kano model.

Feature (attribute) evaluation is an important component of many machine learning tasks, e.g. feature subset selection, constructive induction, decision and regression tree learning. Scores assigned to attributes during evaluation also provide important information to the domain expert trying to get an insight into the problem domain. We are interested in a subclass of feature evaluation, namely the evaluation of conditionally strongly dependent ordinal attributes where each of the individual attribute's values may depend on other attributes in a different way. The problem of feature (attribute) evaluation has received a lot of attention in literature. There are several measures for the evaluation of attributes' quality. For classification problems the most popular are e.g. Gini index (Breiman et al., 1984), gain ratio (Quinlan, 1993), MDL (Kononenko, 1995), and ReliefF (Robnik-Šikonja & Kononenko, 2003). The first three are impurity based and measure quality of attribute according to the purity of class value distribution after the split on the values of that attribute. They evaluate each attribute separately, are not aware of the ordering of the attribute's values and cannot provide useful information for each individual value of the attribute. ReliefF, on the other hand, is context sensitive (by measuring how the attribute separates similar instances) and could be adapted to handle ordered attributes (by changing the definition of its similarity measure), but cannot provide information for each value separately and does not differentiate between the positive and negative changes of the attribute and their impact on the class value.

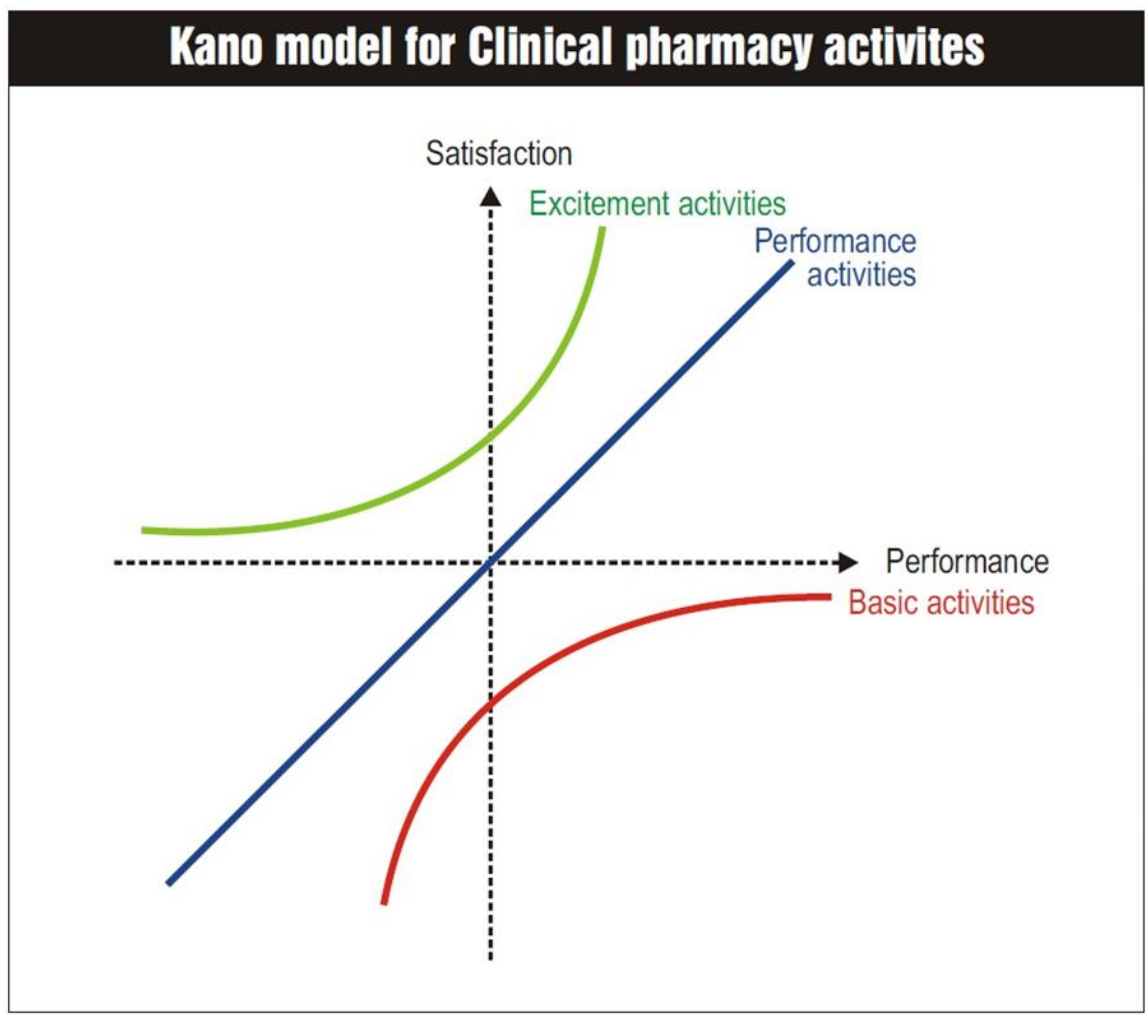

**Figure 1: The Kano model for clinical pharmacy activities**

The goal of feature analysis in our research is manifold:

1. identify features which present the most important competences and influence the overall (dis)satisfaction most,

2. identify types of features; based on total quality management (Berger et al., 1993; Kano et al., 1984) we differentiate between three types of important features:

   a) Basic features are taken for granted by users. High score in these features does not significantly increase the overall satisfaction, while a low score usually causes dissatisfaction; an example of such feature would be ability to provide information about accessibility of medications, which is taken for granted and will not increase satisfaction but will almost certainly cause dissatisfaction if not provided by clinical pharmacists.

   b) Performance features are important features not taken for granted; they usually have a positive correlation with overall satisfaction: the higher the score, the bigger the effect on the overall satisfaction. An example would be personal involvement and effort of the personnel which will correlate positively with satisfaction of the users.

    c)    Excitement features usually describes properties of product/service which are normally not very important to the users, but can cause excitement (and boost in satisfaction) if the score is very high. An example would be participating in clinical studies which will be an important feature for some physicians and nurses.

3.    Identify those attribute values (thresholds) which have a positive/negative impact on overall satisfaction.

There are many different feature evaluation algorithms in data mining which can evaluate, rank and/or select the most informative features (goal 1). Other goals (2 and 3) are tackled by the OrdEval algorithm but remain mostly untouched by other work in machine learning and data mining.

The OrdEval algorithm can be used for analysis of any data where the dependent variable has ordered values, meaning that it is also suitable for surveys where answers are given in the graded manner. The methodology uses conditional probabilities called 'reinforcement factors' as they approximate the upward and downward reinforcement effect the particular feature value has on the dependent attribute. For each value of the feature we obtain estimates of two conditional probabilities: the probability that the response value increases given the increase of the feature value (upward reinforcement), and the probability that the response value decreases given the decrease of the feature value (downward reinforcement). To take the context of other features into account, these probabilities are computed in the local context, from the most similar instances. The visualization of these factors with box-plots gives clear clues about the role of each feature, the importance of each value and the threshold values. To understand the idea of the OrdEval algorithm, the feature should not be treated as a whole. Rather than that, we shall observe the effect a single value of the feature may have.

We also compute confidence intervals for each reinforcement factor. Since we cannot assume any parametric distribution and have to take the context of a similar respondent into account we construct bootstrap estimates and form confidence intervals based on them (Efron & Tibshirani, 1993). We plot obtained random reinforcement factors based on bootstrap sampling with a box-and-whiskers plot: the box is constructed from the 1st and 3rd quartile, middle line is median, while the whiskers are $100\alpha/2$ and $100(1- \alpha)/2$ percentiles (e.g. 2.5 and 97.5 percentiles) giving the borders of confidence interval (e.g., 95% confidence interval). The box-and-whiskers showing confidence intervals are put above the box-plots showing reinforcement factors (Robnik-Šikonja et al., 2009).

We used the OrdEval algorithm for evaluating the importance of a particular clinical pharmacy activity and its influence on the overall satisfaction with clinical pharmacy services. The Algorithm deals independently with each value giving us insight into a particular activity value, its positive or negative influence, the strength of the influence, and the information whether the influence is significant or not.

## 3.2 Classification models

In the introduction of clinical pharmacy comprehensibility of the models is an important factor. Since we have only a limited amount of data available we opted for simple and comprehensible models and selected naïve Bayesian classifier and decision trees. To check whether we lose some predictive accuracy compared to more advanced methods, we also tried random forests (Breiman, 2001), which is known for its robust performance, but did not observe any advantages over the simple methods.

The naïve Bayes classifier is based on a presumption of ideal independency of attributes, and classifies them according to their important features, minimizing the probability of misclassification. It also works on small training sets which gives it its main advantage (Gupta & Suma, 2013). It has been used successfully for attitude and behavior prediction, and we also expected it to be a successful tool in constructing the prediction model for introduction of clinical pharmacy in the hospital.

Decision trees are branched structures of decision sets which generate classification rules for data sets. The decision trees are part of a standard toolbox of numerous analytical software solutions.

# 4 Understanding of important factors for clinical pharmacy

The aim of the data analysis was to discover the most influential clinical pharmacy activities regarding the total satisfaction with clinical pharmacy service as a whole. The OrdEval algorithm was used for processing of the survey data. Based on the findings we constructed a prediction model for the introduction of clinical pharmacy services on the wards, where such cooperation has not yet been established. We aimed to identify the wards with the highest probability of successful cooperation with a clinical pharmacist. We used a Naïve Bayes and decision tree classifiers to construct a prediction model. The model was then used to compute the estimated overall satisfaction value for each of the wards where cooperation has not yet been established.

## 4.1 Results of the OrdEval

Using the OrdEval algorithm we performed an analysis and a visualization of the results for each attribute (question), taking into account all the answers to a particular question. Attribute importance scores (reinforcement factors) computed by the OrdEval algorithm are presented on Figure 2. The length of the bar reflects the probability of the influence of a particular attribute (question) on the overall satisfaction value. Red bars (right from the median line) represent positive influence, which means that in case of increasing the value for a particular question the overall satisfaction value would increase. The contrary holds for blue bars: decreasing of the value of a particular question would decrease the overall satisfaction value.

We can notice a statistically significant influence in two questions (the length of the bar reaching beyond the box-and-whiskers). In the case of decreasing of the values for questions 26 and 38, the overall satisfaction value could decrease. We assume that poor statistical significance for other questions is partly the result of a relatively small sample we had. Therefore we also decided to take into account some other questions with substantial influence (the longest bars) reaching at least beyond the box, which means that the probability of the reinforcement factor lies within the fourth quartile.

We can see that questions 21, 26 and 38 have the longest red and blue bars which means that changing the value for these questions upwards or downwards would increase/decrease the overall satisfaction value. This kind of attributes are designated as **performance** attributes, where quality and/or quantity of activities are the key determinants for the customer's satisfaction or dissatisfaction (Deng et al., 2008) .

In questions 14, 17, 28 and 33 we can notice a substantial length of blue bars, which means that decreasing the value for a particular question would decrease the overall satisfaction value while increasing the particular value would not influence the overall satisfaction value. In this case we talk about **basic** attributes or dissatisfiers because they can only cause dissatisfaction when not observed at an adequate level. Besides performance and basic attributes, the Kano model defines **excitement** attributes (satisfiers) which can only increase the overall satisfaction value when observed at sufficient level but having no influence when they are not observed (Deng et al., 2008).

Among all the questions, question 38 has the most noticeable positive and negative impacts. It is pertaining to the provision of all kind of information about medicines. Providing of objective quality information about medicines is one of the most frequent activities of clinical pharmacists and it has also been proven to be successful elsewhere (Chevalier & Neville, 2011; Kjeldsen, Jensen, & Jensen, 2011). Besides the negative reinforcement, the question 38 also has the strongest upwards reinforcement factor which is however not statistically significant, since its downwards reinforcement is stronger.

When taking into account all the answers for each question, there were no excitement attributes identified by the OrdEval algorithm in our dataset.

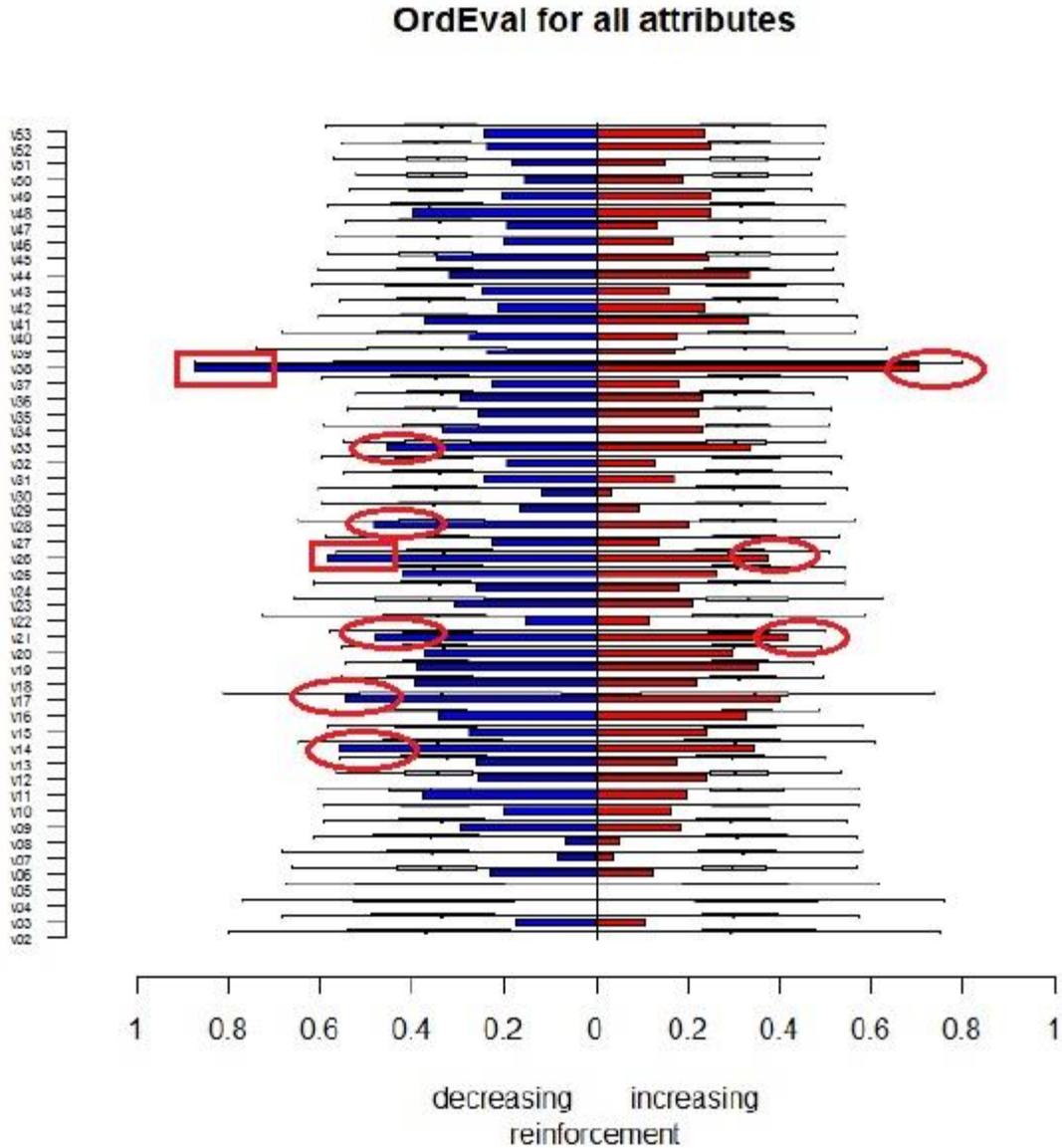

**Figure 2: OrdEval algorithm values for all attributes**

As the OrdEval algorithm also enables feature level detailed data analysis and visualization, taking into account a value change for a certain feature, we computed and visualized all the questions and their reinforcement factors at each value. The length of the bars reflects the strength of the influence of the change at a certain value. For each value also the 95% confidence interval is computed and visualized with box-and-whiskers plot above each bar.

The bars reaching beyond the box-and-whiskers plot depict statistically significant reinforcement factors for each upwards or downwards value change. Besides seven statistically significant reinforcement factors we also analyzed the questions with reinforcement factor at least 0.6. We assumed the questions with reinforcement probabilities lower than 0.6 have a neglecting influence.

In a detailed analysis, we identified six questions, 9, 19, 21, 26, 33 and 38, where decreasing of a value could decrease the overall satisfaction value while increasing of a value would increase the overall satisfaction. These questions are therefore assigned as **performance** attributes. The questions and the OrdEval visualization of the reinforcement factors for performance attributes are shown in Figure 3.

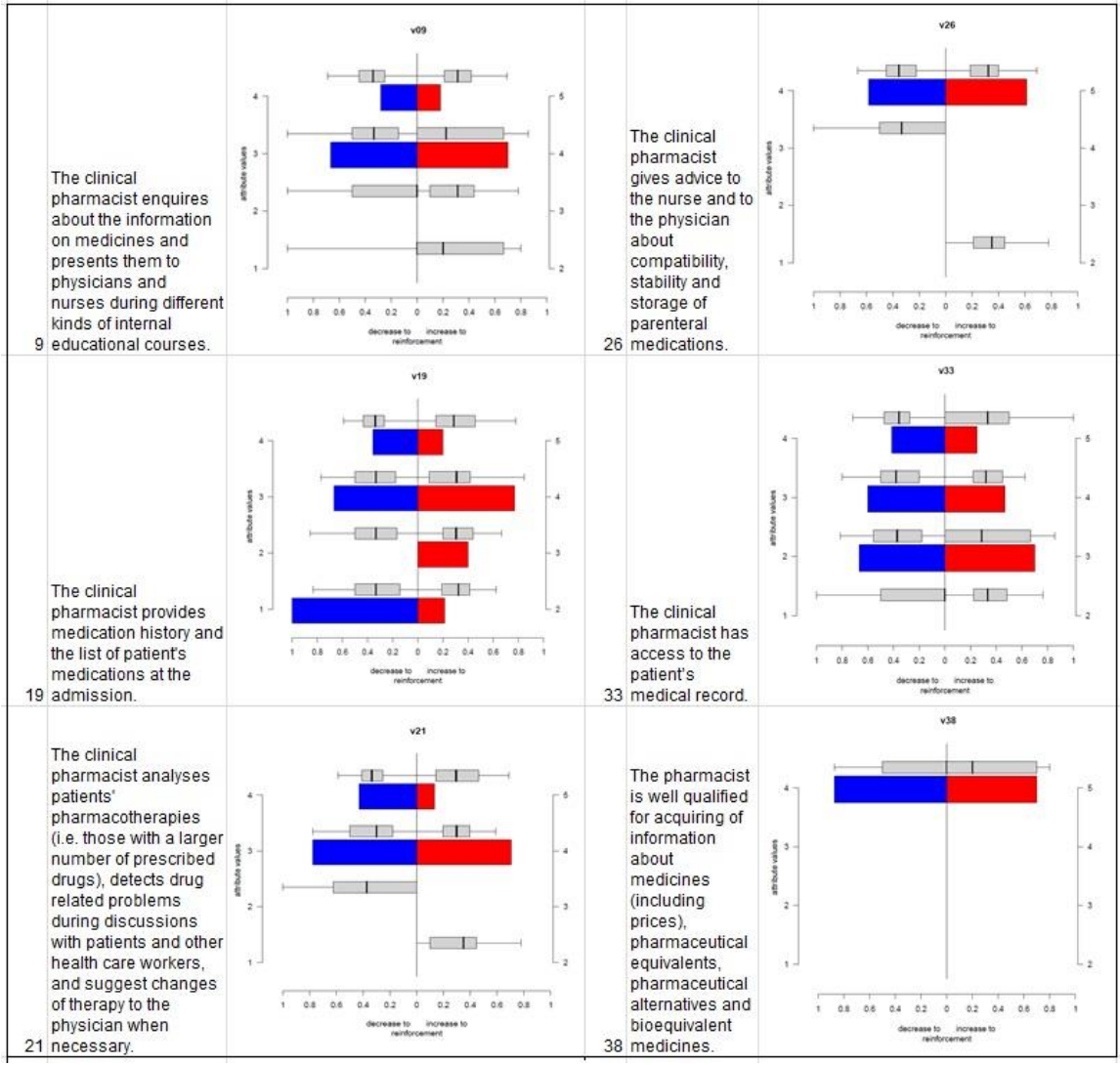

**Figure 3: A visualization of the OrdEval algorithm for questions where increasing/decreasing of a certain value could increase/decrease the overall satisfaction value.**

The question 33 deals with the competence of pharmacists rather than their activity. There are some other questions like this in the survey. Performance of an activity has a direct impact on its quality and consequently on the satisfaction. On the other hand this is not true for the competence. Rather we can say that performance is a function of competence. In future it would be reasonable to include into the satisfaction survey questionnaire only questions concerning activities and not competencies.

The analysis of the influences of other questions gives us very useful information for directing the clinical pharmacists' activities. On the wards where the cooperation of a clinical pharmacist was established, pharmacists mostly perform the activities described in questions 9, 19, 21, 26 and 38. Based on informal contacts with the medical and nursing staff members we realized that satisfaction with the cooperation with a clinical pharmacist depends on quantity, namely on the amount of time that pharmacists are able to dedicate to clinical pharmacy services and the amount of work done on the ward. Some of the ward chief physicians express their demands very clearly saying they don't want clinical pharmacy to only be a boutique service for selected patients. Rather than that, they want this service for all of their patients.

In questions 19 and 38, only downwards reinforcement factors are statistically significant. Therefore we can conclude that these two activities have a tendency to become basic (dissatisfiers) attributes where decrease of the value would decrease the overall satisfaction while increasing the value wouldn't have a considerable influence.

With question 21 we can notice a statistically significant influence on the overall satisfaction value when decreasing the value from 3 to 2 but also when increasing the value from 3 to 4. Considering the mean scores for physicians and nurses in the survey, 4.2 and 3.0 respectively, the result of the OrdEval algorithm is expected. A statistically significant influence of decreasing the value from 2 to 1 for question 19 can be noticed. The mean scores of physicians and nurses in the survey were 3.5 and 2.9 respectively, which indicates relatively low expectations. As the pharmacists' mean score for this question was 4.8 we can anticipate that the decrease from 2 to 1 is not really likely to happen. In that case we can assign certain importance also to the upwards reinforcement factor when changing the value from 3 to 4, although it is statistically not significant. For that reason we included the question 19 into the performance attributes group. On the basis of the results obtained with the OrdEval algorithm we can assume that this activity is going to become a basic one. It was shown to be one of the most valuable activities of clinical pharmacists in (Kaboli et al., 2006; Knez et al., 2011; Režonja et al., 2010).

In the visualization of the results for questions 14, 15, 16, 17, 32, 35, 36, 41, 44 and 46 we notice a substantial asymmetry of the reinforcement factors in the upwards direction. In case of increase of a value for these questions the overall satisfaction value would increase. Activities described in these questions therefore belong to a group of excitement attributes. Their description and visualization are shown in Figure 4.

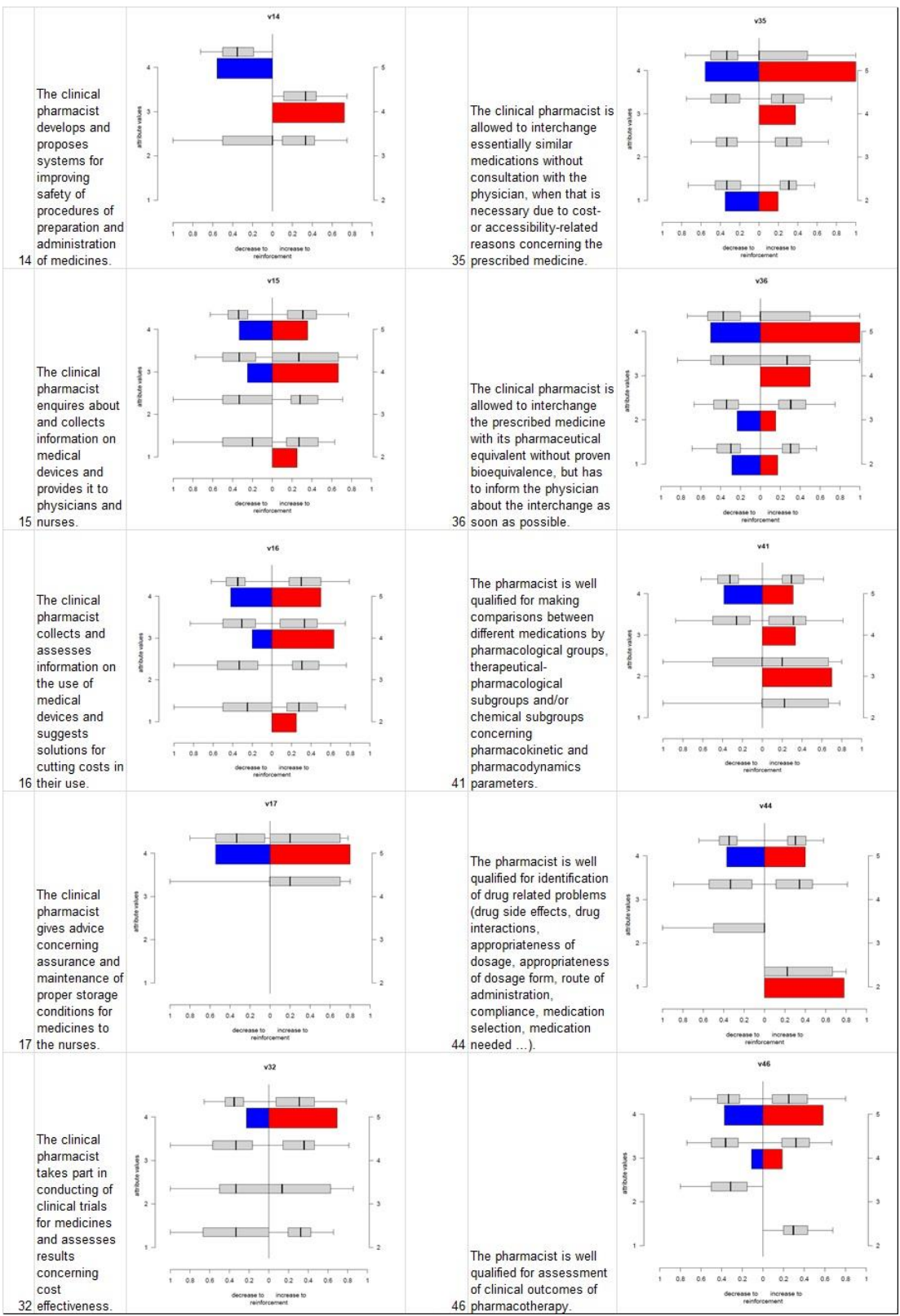

**Figure 4: A Visualization of the OrdEval algorithm for questions where increasing of a certain value could increase the overall satisfaction value.**

The questions 35 and 36 describe pharmacists' competences and we cannot use them for the evaluation of the satisfaction with the clinical pharmacy service (same as for the question 33). Anyway, the results of the OrdEval algorithm give us important information for the planning and direction of clinical pharmacy services. Despite low median scores for these two questions from physicians and nurses we can expect an increase in the overall satisfaction when increasing scores for these two questions.

According to the results given by the OrdEval algorithm, question 17 is an excitement attribute, too. High mean scores reached by physicians (4.8) and nurses (4.5) in the survey could lead us to a false conclusion that this activity is basic, however the OrdEval algorithm shows us an opportunity to increase the overall satisfaction value. Reinforcement factors for other questions in this group are high as well and are meaningful and useful in planning of activities. Although statistically not significant (possibly due to a small sample) they show us the activities with considerable potential of increasing the overall satisfaction value.

The only question where analysis shows a substantial downwards reinforcement factor is question 11. Decreasing the value from 3 to 2 would have statistically significant decreasing impact to the overall satisfaction value. Consequently we place this activity into the **basic** type of attributes.

While analyzing the detailed attribute results of the OrdEval algorithm we detected the same set of important questions as when considering all scores of a certain question. The only exception is question 28 which showed too little impact (reinforcement $< 0.6$) in detailed analysis. Besides these, the detailed analysis revealed some other important questions. However, the most important finding of the comparison of the two levels of the analysis is that in some questions the direction of influence has changed when examining detailed results. For example, questions 14 and 17 would be in the basic group of attributes when considering average over all values, while value level analysis displays the excitement feature of these two attributes.

Furthermore we can notice that for some questions the reinforcement factors are extremely low (value $< 0.4$) and some values even do not occur in the visualization of the collected answers (Fig. 2, questions 7,8 and 3,4). Some of these questions have similar context and it would be reasonable to merge them into a common-meaning question in the future. This analysis unveils another application of the OrdEval algorithm, namely it can be used for testing and validating survey questions during the development phase of a questionnaires'.

We noticed in some cases that the overall satisfaction with the cooperation with clinical pharmacists depends not only on the services they provide but also on their personality characteristics where a proactive approach is desired. With future evaluations of satisfaction it would be necessary to consider also the personal characteristics of pharmacists.

There are at least two notable advantages of the OrdEval algorithm for managerial analysis. First, it provides user-friendly visualization supporting decision making process, and second, its two-level structure enables a detailed insight into the direction and strength of the influence of particular values and thresholds.

## 4.2 Prediction models

In order to construct the prediction model we used the 10-fold cross validation method, which is known to be suitable for small samples (Molinaro, Simon, & Pfeiffer, 2005). We used Orange Canvas environment (Demšar et al., 2013). Among different classifiers tested, Naive Bayes, classification trees, and random forests, gave us the best classification accuracies, 57%, 55% and 50% respectively. These classification accuracies are better than 40.9%, which is a prior probability for the most frequent answer 5.

If we assume predictions differing only for one value are still acceptable, the classification accuracies of the classification tree and naive Bayes are 77.3% and 72.7%, respectively. These classification accuracies are good enough and we can recommend these two classifiers to be used for prediction purposes in the future.

The constructed prediction model forecasts the highest value (5) for four clinical departments, a value of 4 for six clinical departments and a value of 3 or less for other clinical departments. From a practical point of view it is the most important for a prediction model to correctly predict the highest values. After the analysis we introduced clinical pharmacists to some of the clinical departments with the highest predicted value. We currently have a very positive feedback in all departments, although the satisfaction has not been measured objectively yet. We can conclude that in spite of using the analytical tools known to be suitable for small samples, the sample size remains a limiting factor for our study, which has to be considered while interpreting the prediction model results.

When introducing further clinical pharmacy services, we will have to take into account that the predicted values hold to some degree for the heads of the department but not necessarily for the whole department and for all the physicians.

# 5  Conclusions

We adapted a general methodology for analysis of ordered data to the specifics of human resources management in hospital. We were able to obtain useful managerially relevant information from a relatively small sample of highly relevant respondents. We showed how the OrdEval algorithm exploits the information hidden in the ordering of class and attribute values and their inherent correlation. The algorithm can handle ordered attributes and ordered classes, is aware of the information the ordering contains, and is able to handle each value of the attribute separately. The provided output can be effectively visualized and complemented with confidence intervals for reinforcement factors. The visualizations turned out useful in our clinical pharmacy research case study.

Contrary to other survey analysis algorithms, using the OrdEval algorithm we could get a deep insight into the opportunities to influence the satisfaction with clinical pharmacy services. OrdEval served us as a tool to reveal the crucial attributes which, when decreasing, decrease the overall satisfaction value, and increase the overall satisfaction value when increasing. We were able to detect basic, performance, and excitement attributes expressed in our survey questions, which was important for planning and managing of the introduction of new clinical pharmacy services in clinical departments of the hospital.

While we had good reasons to limit the size of the sample to the most important decision makers, from analytical point of view it would be reasonable to take a larger sample. A future survey involving all the physicians and nurses would be highly welcome and would allow construction of a new prediction model. Among prediction models constructed on existing data, we selected the model with the highest number of correctly predicted high values. As we introduce clinical pharmacists to a limited number of departments, from practical point of view, they will be only introduced to the departments with the high prediction value. The constructed prediction model was found to be successful and useful by the head of the pharmacy who was responsible for human resources allocation. In the future an objective measure of success has to be used to validate the generated model to avoid a possible bias introduced by subjective judgment of the head of pharmacy.

Finally, the survey data sets are important sources of data and their successful analysis is of key importance for informed policy-decision making. We showed that the OrdEval can be efficient in this respect.

## *Acknowledgements*

Andreja Čufar was supported by the University of Ljubljana innovative sheme for PhD study. Aleš Mrhar and Marko Robnik-Šikonja were supported by the Slovenian Research Agency, ARRS, through research programmes P1-0189 (B) and P2-0209, respectively.